\documentclass[letterpaper, 10 pt, conference]{ieeeconf}  

\usepackage{times}
\usepackage[numbers]{natbib}
\usepackage{multicol}
\usepackage[bookmarks=true]{hyperref}

\usepackage{lipsum}
\usepackage{comment}
\usepackage{balance}
\usepackage{graphicx}
\usepackage[table]{xcolor}
\usepackage{caption}
\usepackage{listings}

\usepackage{seqsplit}

\usepackage{amssymb}
\usepackage{amsmath} 
\usepackage{multirow}

\newcommand{\tabincell}[2]{\begin{tabular}{@{}#1@{}}#2\end{tabular}}

\IEEEoverridecommandlockouts                              

\usepackage{tabularx} 

\title{\Large \bf
Optimizing Active Perception for Learning Simultaneous Viewpoint Selection and Manipulation with Diffusion Policy
}

\author{Xiatao Sun, Francis Fan$^{*}$, Yinxing Chen$^{*}$, Daniel Rakita
\thanks{$^{*}$ Equal contribution}
\thanks{All authors are with the Department of Computer Science, Yale University, New Haven, CT 06520, USA
        {\tt\small \{xiatao.sun, francis.fan, j.y.chen, daniel.rakita\}@yale.edu}}%
\thanks{This work was supported by Office of Naval Research award N00014-24-1-2124}
}

\begin{document}

\maketitle
\thispagestyle{empty}
\pagestyle{empty}


\begin{abstract}

Robotic manipulation tasks often rely on static cameras for perception, which can limit flexibility, particularly in scenarios like robotic surgery and cluttered environments where mounting static cameras is impractical. Ideally, robots could jointly learn a policy for dynamic viewpoint and manipulation. 
However, dynamic viewpoint control requires additional degrees of freedom and intricate coordination with manipulation, which results in more challenging policy learning than single-arm manipulation. To address this complexity, we propose an integrated learning framework that combines diffusion policy with a novel look-at inverse kinematics solver for active perception. Our framework helps better coordinating between perception and manipulation. It automatically optimizes camera orientation for viewpoint selection, while allowing the policy to focus on essential manipulation and positioning decisions. We demonstrate that our integrated approach achieves superior performance and learning efficiency compared to directly applying diffusion policies to configuration space or end-effector space with various rotation representations.
Further analysis suggests that these performance differences are driven by inherent variations in the high-frequency components across different state-action spaces.

\end{abstract}

\section{Introduction}
\label{sec:introduction}

Robotic manipulation with fixed viewpoints has been extensively studied and deployed in real-world scenarios. Traditionally, viewpoints for manipulation tasks are either static in the workspace or rigidly attached to the manipulation arm, such as with a wrist camera. However, this fixed-viewpoint paradigm is limited in scenarios where mounting static cameras is impractical or provides limited perception. For instance, in modern robotic surgery, a robotic arm with a mounted camera needs to be controlled alongside other arms with instruments \cite{ettorre_robotic_surgery}. In cluttered environments, a single camera may be obstructed by surrounding objects, while multiple cameras increase computational costs due to the additional video streams. 

\begin{figure}
\centering
\includegraphics[width=\columnwidth]{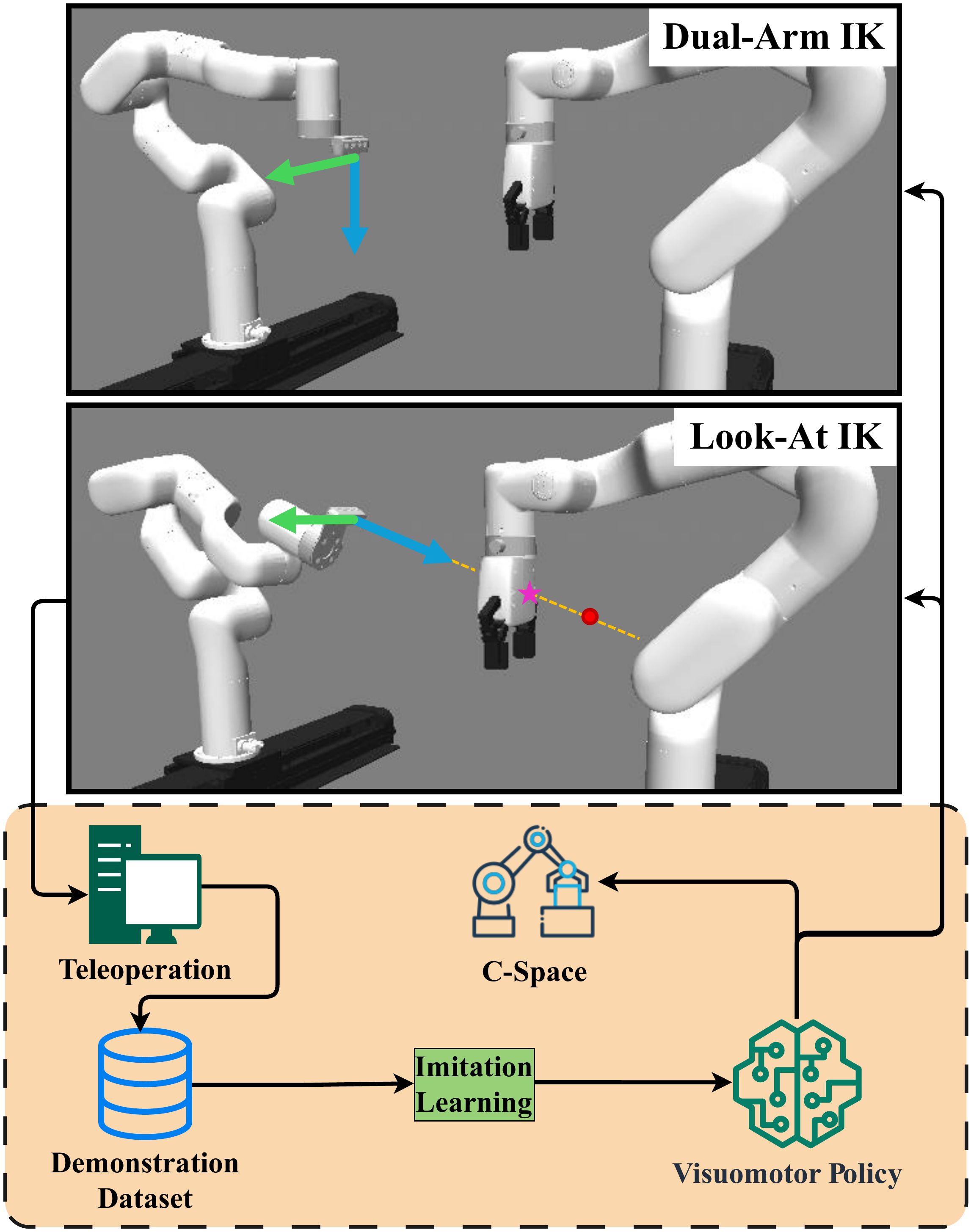}
\caption{An overview of the visuomotor policy learning pipeline and a comparison of the state-action spaces. While policies can be trained in Configuration Space (C-Space) or with a standard Dual-Arm IK, our proposed Look-At IK automatically determines the orientation of the viewpoint arm end-effector. \textcolor[HTML]{47D45A}{Green} and \textcolor[HTML]{0F9ED5}{blue} arrows represent the local $y$ and $z$ axes $\mathbf{w}_y$ and $\mathbf{w}_z$ of the viewpoint end-effector, respectively. The \textcolor[HTML]{C00000}{red} dot indicates a distant point $\mathbf{p_f}$ along the \textcolor[HTML]{FFC000}{orange} line, showing the direction of $\mathbf{w}_z$. The \textcolor[HTML]{E72DC4}{pink} star marks the visual target $\mathbf{p_t}$, set with a fixed offset from $\mathbf{p_m}$.}
\label{fig:ik_comparison}
\vspace{-0.6cm}
\end{figure}

These challenges highlight the need for simultaneous viewpoint selection and manipulation within a robotic system featuring a dynamic camera. Ideally, robots could jointly learn viewpoint selection and manipulation, allowing them to generalize patterns of what, when, and how to move, along with determining when and from where to update their understanding of the environment to complete tasks. However, integrating manipulation with viewpoint selection introduces significant difficulty for visuomotor policy learning due to the increased complexity and intricate coordination required between the camera and the gripper.

To address this challenge, we introduce a novel framework that synergistically combines a visuomotor diffusion policy with a specialized look-at Inverse Kinematics (IK) solver, as illustrated in Fig. \ref{fig:ik_comparison}. Our look-at IK solver, detailed in \S\ref{sec:look_at_ik}, decouples the learning problem by programmatically optimizing the camera's orientation to maintain focus on the manipulation target. This approach defines the look-at space, which is a new, reduced end-effector space that significantly simplifies the learning task for the policy. By handling the complex camera orientation automatically, our framework allows the diffusion policy to focus on the essential Degrees of Freedom (DOF) for the task, leading to improved training efficiency and superior performance. To validate the effectiveness of our framework, we demonstrate its performance against policies trained in more conventional state-action spaces, including the configuration space,  full end-effector space and its variants with different rotation representations.

To facilitate evaluation, we developed new simulation environments for dual-arm active perception systems. Our experiments in both simulation and the real world demonstrate that our integrated framework achieves a significantly higher task success rate and better training efficiency (\S\ref{sec:simulation_experiments}, \S\ref{sec:real_world_experiments}). Crucially, our analysis reveals that the performance gains are not solely attributable to the reduced dimensionality, which is a common assumption in policy learning \cite{hamaya_dim, lonvcarevic_dim, abu_dim}. Instead, we find a strong correlation between policy performance and the prevalence of High-Frequency Components (HFC) within the state-action space (\S\ref{sec:frequency_analysis}). Our proposed look-at space exhibits lower HFC, making the learning problem more tractable for the diffusion model architecture. This insight provides a deeper understanding of the factors affecting visuomotor policy performance in complex coordination tasks. In summary, our contributions are:

\begin{itemize}
    \item A novel framework that integrates a diffusion policy with a look-at IK solver to automate optimal viewpoint orientation, significantly improving performance and training efficiency for manipulation tasks with active perception.
    \item New simulation environments for developing and evaluating dual-arm active perception and manipulation systems.
    \item An analysis demonstrating that the performance of visuomotor policies is strongly correlated with the HFC of the state-action space, offering a new perspective beyond just dimensionality.
    \item Open-source implementation of our framework to facilitate future research on viewpoint selection and manipulation.
    \footnote{\href{https://github.com/Apollo-Lab-Yale/spaces_comparative_study}{https://github.com/Apollo-Lab-Yale/spaces\_comparative\_study}}
\end{itemize}

\section{Related Works}
\label{sec:related_works}

\subsection{Robot Learning and Diffusion Models}

Learning-based methods have been successfully applied to various robotic platforms, such as quadrotors \cite{wu_quadrotor_il}, autonomous driving \cite{sun_il_benchmark, zhou_ad}, and robotic arms \cite{wang_robo_learning, kim_il_arm}. In particular, research on Imitation Learning (IL) has focused on improving the training loop for better data efficiency and quality, such as incorporating self-supervision \cite{torabi_bco} or data aggregation \cite{sun_megadagger}. These methods, while enhancing the IL training loop, often use simple neural network architectures with a limited number of parameters.

With the rise of large generative models, recent advances in IL have integrated more sophisticated network architectures, including transformer-based models \cite{shafiullah_il_transformer, wang_il_transformer} and diffusion-based policies \cite{chi_diffusion_policy, zhang_diffusion_dagger}. Diffusion policies, inspired by generative models for images and videos that employ a denoising process \cite{ho_ddpm, peebles_dit}, typically utilize Convolutional Neural Networks (CNNs), such as UNet \cite{ronneberger_unet}, or Transformers \cite{vaswani_transformer} as their backbone. These models have shown significant potential and have been successfully applied to various robotic manipulation tasks \cite{scheikl_diff_deform, li_diff_obj_centric}. 

Concurrent work has also begun to apply diffusion policies to manipulation with active perception.  \citet{xiong2025vision} utilized an end-to-end diffusion policy that directly operates in the full end-effector space to learn active perception and manipulation. In contrast, our framework introduces the look-at IK solver as a structural prior to explicitly decouple camera orientation from the learned policy, which significantly improves both performance and training efficiency.

\subsection{Viewpoint Selection and Manipulation}

Viewpoint selection, or active perception \cite{bajcsy_active_perception}, has been extensively studied on mobile robotic platforms using both optimization-based \cite{falanga_ap_opt, spasojevic_ap_opt} and learning-based methods \cite{tordesillas_ap_learning, bartolomei_ap_learning}. However, its application to robotic manipulation, especially for dual-arm systems with high DOFs, remains limited.

Research has shown that viewpoint selection can improve the success rate of manipulation tasks in cluttered and occluded environments \cite{gualtieri_vp_cluttered, kahn_vp_occluded}. Most optimization-based planners with dynamic viewpoint select the next best view based on metrics of information gain, such as the number of revealed occluded voxels \cite{breyer_nbv}, or the reduction in grasp quality uncertainty \cite{morrison_nbv}. Recent work has started applying learning to viewpoint selection and manipulation. For instance, Saito et al. \cite{saito_rnn_vp} used a recurrent neural network for a 6-DOF manipulation scenario, though they decoupled viewpoint selection and manipulation into separate phases. Lv et al. \cite{lv_sam_rl} utilized model-based reinforcement learning to train a policy for dual 7-DOF arms, showing that learning-based viewpoint selection can enhance task execution compared to static viewpoints. However, this work is limited to short-horizon tasks and without any simplification on the state-action space from the perspective of kinematics.

\section{Technical Overview}
\label{sec:technical_overview}

\subsection{Problem Formulation}

Our proposed framework employs IL to train a visuomotor policy for a 17-DOF dual-arm robotic system shown in Fig. \ref{fig:robotic_system_overview}. This process requires a demonstration dataset $\mathcal{D} = \{r_n\}_{n=1}^{N}$, which contains $N$ demonstration rollouts $r_n$. Each rollout is a sequence of observation-action pairs $(\mathbf{O}_t, \mathbf{A}_t)$ at each timestep $t$. This dataset is collected by a human expert with an intrinsic policy $\pi^*$. The goal of imitation learning is to mimic $\pi^*$ by learning a policy $\pi$ using $\mathcal{D}$ that maps observations $\mathbf{O}_t$ to actions $\mathbf{A}_t$:
$$
\pi: \mathbf{O}_t \rightarrow \mathbf{A}_t
$$

In our framework, $\mathbf{O}_t$ consists of $\mathbf{v}_t$, the RGB image from a camera on the viewpoint arm, and $\mathbf{s}_t$, the state of the robot in a space for policy learning.

\subsection{State-Action Spaces}

A key design choice of our framework is the look-at space, which simplifies the learning problem. By programmatically optimizing the camera's orientation, it decouples viewpoint rotation from the policy's action space, allowing the policy to focus on essential manipulation and positioning tasks.

To rigorously validate this approach, we evaluate our method against two standard baselines: Configuration Space and End-Effector Space. These serve as strong points of comparison due to their use in diffusion policies for manipulation \cite{chi_diffusion_policy, seo2024presto}, including concurrent work on active perception \cite{xiong2025vision}. For a comprehensive analysis, we also evaluate variants of the end-effector and look-at spaces using different rotation representations.

This subsection details these spaces in order of increasing abstraction, from the robot's native Configuration Space, to the task-centric End-Effector Space, and concluding with our proposed Look-At Space.

\begin{figure}
\centering
\includegraphics[width=\columnwidth]{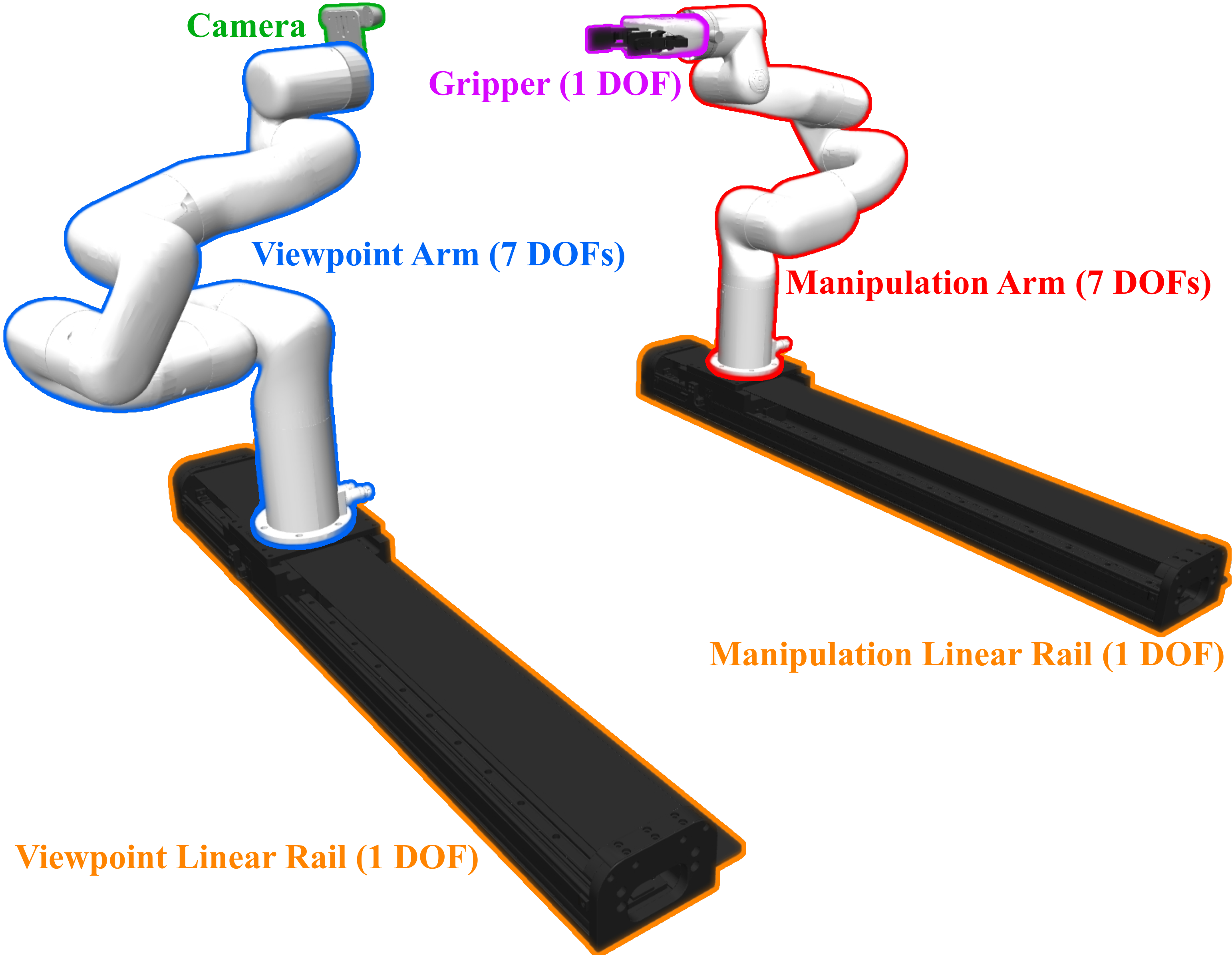}
\caption{Illustration of the 17-DOF dual-arm robotic system used in this paper. The system comprises two 7-DOF robotic arms, each mounted on a linear rail with 1 DOF. The viewpoint arm is equipped with a camera, while the manipulation arm features a gripper with an additional 1 DOF.}
\label{fig:robotic_system_overview}
\vspace{-0.6cm}
\end{figure}

\textbf{Configuration Space (C-Space)} is the foundational representation of a robot's state, directly characterizing its joint angles. As a multidimensional space where each dimension corresponds to a DOF, $\mathcal{C}$ encompasses all possible configurations of the system. For our dual-arm robot:
$$
\mathcal{C} = \{ \mathbf{c} \in \mathbb{R}^{17} \mid c_i^{\text{min}} \leq c_i \leq c_i^{\text{max}}, \forall i \}
$$
where $\mathbf{c}$ is the configuration vector constrained by the minimum and maximum values of each motor of the 17-DOF system, as illustrated in Fig. \ref{fig:robotic_system_overview}.

\textbf{End-Effector Space} offers a more intuitive, task-oriented representation by defining the positions and orientations of the end-effectors in the workspace. States and actions in this space are mapped back to C-Space using a standard IK solver. For our dual-arm system, the end-effector space $\mathcal{E} = \mathcal{E}_m \times \mathcal{E}_v$ combines the manipulation arm end-effector space $\mathcal{E}_m$ and the viewpoint arm end-effector space $\mathcal{E}_v$. The end-effector vector of single arm consists of a position vector $\mathbf{p} = [x, y, z]^T \in \mathbb{R}^{3}$ and a rotation representation. We consider three rotation representations: Euler angles $\mathbf{R}_{\text{E}} \in \mathbb{R}^{3}$, unit quaternions $\mathbf{R}_{\text{Q}} \in \mathbb{H}_1$, and axis-angle $\mathbf{R}_{\text{AA}} \in (\mathbb{R}^3, \mathbb{R})$. For Euler angles, $\mathbf{R}_{\text{E}} = [\alpha, \beta, \gamma]^T$, where $\alpha$, $\beta$, and $\gamma$ denote roll, pitch, and yaw angles. For unit quaternions, $\mathbf{R}_{\text{Q}} = [q_0, q_1, q_2, q_3]^T$, with $q_0$ as the scalar part, $[q_1, q_2, q_3]^T$ as the vector part, and $q_0^2 + q_1^2 + q_2^2 + q_3^2 = 1$. For axis-angle, $\mathbf{R}_{\text{AA}} = (\mathbf{u}, \theta)$, where $\mathbf{u} = [u_x, u_y, u_z]^T$ is a unit vector representing the rotation axis, and $\theta$ is the rotation angle in radians.

Based on the single-arm end-effector representation, we consider three variants of the dual arm end effector space, including the Euler space of the end effector $\mathcal{E}_{\text{E}} = \{ [\mathbf{p}_m, \mathbf{R}_{\text{E},m}, \mathbf{p}_v, \mathbf{R}_{\text{E},v}] \}$, the quaternion space of the end effector $\mathcal{E}_{\text{Q}} = \{ [\mathbf{p}_m, \mathbf{R}_{\text{Q},m}, \mathbf{p}_v, \mathbf{R}_{\text{Q},v}] \}$, and the axes angle space of the end effector $\mathcal{E}_{\text{AA}} = \{ [\mathbf{p}_m, \mathbf{R}_{\text{AA},m}, \mathbf{p}_v, \mathbf{R}_{\text{AA},v}] \}$. $\mathbf{p}_m$ and $\mathbf{p}_v$ are the respective positions of the manipulation and viewpoint end-effectors, $\mathbf{R}_{\text{E},m}$ and $\mathbf{R}_{\text{E},v}$ are the respective Euler angles of the manipulation and viewpoint end-effectors, $\mathbf{R}_{\text{Q},m}$ and $\mathbf{R}_{\text{Q},v}$ are the respective quaternions of the manipulation and viewpoint end-effectors, and $\mathbf{R}_{\text{AA},m}$ and $\mathbf{R}_{\text{AA},v}$ are the respective axis-angle representations of the manipulation and viewpoint end-effectors.

During policy learning from the end-effector space, all elements are treated as real numbers to facilitate fitting the neural network. Therefore, concatenating with the gripper configuration, policy learning with $\mathcal{E}_{\text{E}}$, $\mathcal{E}_{\text{Q}}$, and $\mathcal{E}_{\text{AA}}$ is actually in $\mathbb{R}^{13}$, $\mathbb{R}^{15}$, and $\mathbb{R}^{15}$ respectively.

\textbf{Look-At Space} is a principled simplification of the end-effector space for active perception and manipulation scenarios by excluding the orientation of the viewpoint arm, which is automatically determined (see \textsection IV-B) \cite{rakita_lookatik}. Similar to the end-effector space, we consider three variants of the look-at space: Look-At Euler Space $\mathcal{L}_{\text{E}} = \{ [\mathbf{p}_m, \mathbf{R}_{\text{E},m}, \mathbf{p}_v] \}$, Look-At Quaternion Space $\mathcal{L}_{\text{Q}} = \{ [\mathbf{p}_m, \mathbf{R}_{\text{Q},m}, \mathbf{p}_v] \} $, and Look-At Axis-Angle Space $\mathcal{L}_{\text{AA}} = \{ [\mathbf{p}_m, \mathbf{R}_{\text{AA},m}, \mathbf{p}_v] \}$ after excluding $\mathbf{R}_{\text{E},v}$, $\mathbf{R}_{\text{Q},v}$, or $\mathbf{R}_{\text{AA},v}$.

Similar to the end-effector space, policy learning with $\mathcal{L}_{\text{E}}$, $\mathcal{L}_{\text{Q}}$, $\mathcal{L}_{\text{AA}}$ is in $\mathbb{R}^{10}$, $\mathbb{R}^{11}$, and $\mathbb{R}^{11}$ respectively.

\subsection{Diffusion Policy}
Our framework uses a diffusion policy \cite{chi_diffusion_policy} as the visuomotor component, which is trained by IL within the state-action spaces described previously. We use U-Net \cite{ronneberger_unet} as the backbone for the diffusion policy. The diffusion policy employs receding horizon control and models the conditional distribution $p(\mathbf{A}_t \mid \mathbf{O}_t)$. During inference, the diffusion policy starts from Gaussian noise $\mathbf{A}_t^K$, where $K$ is the number of steps for denoising. The denoising process repeatedly utilizes the following equation:
$$
\mathbf{A}_t^{k-1} = \alpha \left( \mathbf{A}_t^k - \gamma \epsilon_\theta (\mathbf{O}_t, \mathbf{A}_t^k, k) + \mathcal{N}(0, \sigma^2, \mathbf{I}) \right)
$$
until the process reaches the output $\mathbf{A}_t^0$ of the policy.

\section{Technical Details}
\label{sec:technical_details}

This section details the IK solvers that map the task-level spaces back to the C-Space. We first describe the standard End-Effector IK, which serves as the baseline and foundation for our method. We then present our proposed Look-At IK solver, which introduces a novel optimization objective specifically designed to automate viewpoint control for active perception with manipulation tasks.

\subsection{End-Effector IK}
\label{sec:dual_arm_ik}

A traditional end-effector IK solver, when applied to our dual-arm system, is designed to convert from any end-effector space variant $\mathcal{E}_{\text{E}}, \mathcal{E}_{\text{Q}}, \mathcal{E}_{\text{AA}}$ to the C-space $\mathcal{C}$. The goal is to find the joint configurations $\mathbf{c}$ that minimize the transformation error between the desired and current positions $\mathbf{p}_{\text{m}}^*, \mathbf{p}_{\text{m}}$ and orientations $\mathbf{R}_{\text{m}}^*, \mathbf{R}_{\text{m}}$ of the manipulation end-effector, and similarly for the viewpoint end-effector.

To achieve this, we utilize Forward Kinematics (FK) to determine the current positions and orientations from $\mathbf{c}$:
$$
\mathbf{p_m}, \mathbf{R_m}, \mathbf{p_v}, \mathbf{R_v} = \text{FK}(\mathbf{c})
$$
In this process, we employ quaternions $\mathbf{R}_{\text{Q}}$ for rotation representation, due to their numerical stability and ability to avoid gimbal lock. The FK results, expressed in quaternion form, are then converted back to the appropriate rotation representations ($\mathbf{R}_{\text{E}}, \mathbf{R}_{\text{Q}}, \mathbf{R}_{\text{AA}}$) for the respective state-action space variants.

The transformation errors are calculated using Euclidean distance for positions and displacement-based distance for quaternions, where the quaternion distance between two quaternions $\mathbf{R}_1$ and $\mathbf{R}_2$ is defined as:
$$
d(\mathbf{R}_1, \mathbf{R}_2) = \min \left( \| \ln (\mathbf{R}_1^{-1} \mathbf{R}_2)^\vee \|_2^2, \| \ln (\mathbf{R}_1^{-1} (-\mathbf{R}_2))^\vee \|_2^2 \right)
$$
which accounts for the antipodal equivalents of quaternions. $ln$ represents the logarithm map that transforms a rotation from the Lie Group to its corresponding Lie Algebra, and $\vee$ is the operator that maps the Lie Algebra element to its corresponding vector in Euclidean space \citep{spatial_modeling}.

The objective function for a single arm is then given by:
$$
J_{\mathcal{E}, i}(\mathbf{p_i}, \mathbf{R_i}, \mathbf{p_i}^*, \mathbf{R_i}^*) = \| \mathbf{p_i} - \mathbf{p_i}^* \|_2^2 + d(\mathbf{R_i}, \mathbf{R_i}^*)
$$

For both arms combined, the objective function is:
\begin{align*}
J_{\mathcal{E}}(\mathbf{c}, \mathbf{p}_{\text{m}}^*, \mathbf{R}_{\text{m}}^*, \mathbf{p}_{\text{v}}^*, \mathbf{R}_{\text{v}}^*) = & J_{\mathcal{E}, \text{m}}(\mathbf{p_m}, \mathbf{R_m}, \mathbf{p}_{\text{m}}^*, \mathbf{R}_{\text{m}}^*) \\
& + J_{\mathcal{E}, \text{v}}(\mathbf{p_v}, \mathbf{R_v}, \mathbf{p}_{\text{v}}^*, \mathbf{R}_{\text{v}}^*)
\end{align*}

The optimal joint configuration $\mathbf{c}^*$ is found by minimizing this combined objective function.

\subsection{Look-At IK}
\label{sec:look_at_ik}

Building on the traditional end-effector IK solver, the look-at IK solver is introduced to further enhance viewpoint selection. It incorporates an additional viewpoint-selection loss, which automatically calculates the orientation of the viewpoint end-effector. This reduces the state-action space to a look-at space variant $\mathcal{L}_{\text{E}}, \mathcal{L}_{\text{Q}}, \mathcal{L}_{\text{AA}}$, which has lower dimensionality compared to the full end-effector space variants $\mathcal{E}_{\text{E}}, \mathcal{E}_{\text{Q}}, \mathcal{E}_{\text{AA}}$.

The look-at IK solver continues to use the FK process for determining the current state of the end-effectors. Although the orientation $\mathbf{R_v}$ of the viewpoint is excluded from the state-action space, it is still necessary to extract the local $y$ axis $\mathbf{w}_y$ and $z$ axis $\mathbf{w}_z$ from $\mathbf{R_v}$ for the optimization process.

The objective function for the manipulation arm remains unchanged:
$$
J_{\mathcal{L}, m}(\mathbf{p_m}, \mathbf{R_m}, \mathbf{p_m}^*, \mathbf{R_m}^*) = \| \mathbf{p_m} - \mathbf{p_m}^* \|_2^2 + d(\mathbf{R_m}, \mathbf{R_m}^*)
$$

For the viewpoint arm, the orientation is determined by minimizing a viewpoint selection objective function, which consists of two main components: a viewpoint orientation objective term and a viewpoint stability objective term.

The viewpoint orientation objective term measures the distance between the visual target $\mathbf{p_t}$ and its projection onto a line segment defined by the viewpoint position $\mathbf{p_v}$ and a distant point $\mathbf{p_f}$, located along the $z$ axis direction $\mathbf{w}_z$:
$$
\mathbf{p_f} = \mathbf{p_v} + \delta \mathbf{w}_z
$$
In practice, we set the distance factor $\delta$ to a large constant (e.g., 999) to ensure $\mathbf{p_f}$ is always sufficiently distant.

To find this distance, the visual target $\mathbf{p_t}$ is projected onto the line segment using the clamped projection function:
$$
\text{proj}_{clamped}(\mathbf{v}, \mathbf{u}) = \min(\max(\text{proj}_{scalar}(\mathbf{v}, \mathbf{u}), 0), 1) \cdot \mathbf{u}
$$

The line segment projection of $\mathbf{p_t}$ is then:
$$
\text{proj}_{line}(\mathbf{p_t}, \mathbf{p_v}, \mathbf{p_f}) = \text{proj}_{clamped}(\mathbf{p_t} - \mathbf{p_v}, \mathbf{p_f} - \mathbf{p_v}) + \mathbf{p_v}
$$

The viewpoint orientation objective function is defined as:
$$
J_{\mathcal{L}, vo}(\mathbf{p_t}, \mathbf{p_v}, \mathbf{p_f}) = \left\| \text{proj}_{line}(\mathbf{p_t}, \mathbf{p_v}, \mathbf{p_f}) - \mathbf{p_t} \right\|_2^2
$$

The viewpoint stability objective term is designed to keep the viewpoint upright, preventing rolling and ensuring consistent orientation. This is calculated as:
$$
J_{\mathcal{L}, vs}(\mathbf{w}_y) = (\mathbf{w}_y \cdot [0,0,1]^T)^2
$$

Combining these components, the complete viewpoint selection objective function is:
\begin{align*}
J_{\mathcal{L}, v}(\mathbf{w}_y, \mathbf{w}_z, \mathbf{p_v}, \mathbf{p_v}^*) = & \| \mathbf{p_v} - \mathbf{p_v}^* \|_2^2 \\
& + J_{\mathcal{L}, vo}(\mathbf{p_t}, \mathbf{p_v}, \mathbf{p_f}) \\
& + J_{\mathcal{L}, vs}(\mathbf{w}_y)
\end{align*}

In this framework, the visual target $\mathbf{p_t}$ is set as the gripper position $\mathbf{p_m}$. Finally, the entire objective function for the look-at IK solver becomes:
\begin{align*}
J_{\mathcal{L}}(\mathbf{c}, \mathbf{p}_{\text{m}}^*, \mathbf{R}_{\text{m}}^*, \mathbf{p}_{\text{v}}^*) = & J_{\mathcal{L}, m}(\mathbf{p_m}, \mathbf{R_m}, \mathbf{p_m}^*, \mathbf{R_m}^*) \\
& + J_{\mathcal{L}, v}(\mathbf{w}_y, \mathbf{w}_z, \mathbf{p_v}, \mathbf{p_v}^*)
\end{align*}

where $\mathbf{c}$ is optimized to minimize this combined objective function.
\section{Evaluation}
\label{sec:evaluation}

\begin{figure}
\centering
\includegraphics[width=\columnwidth]{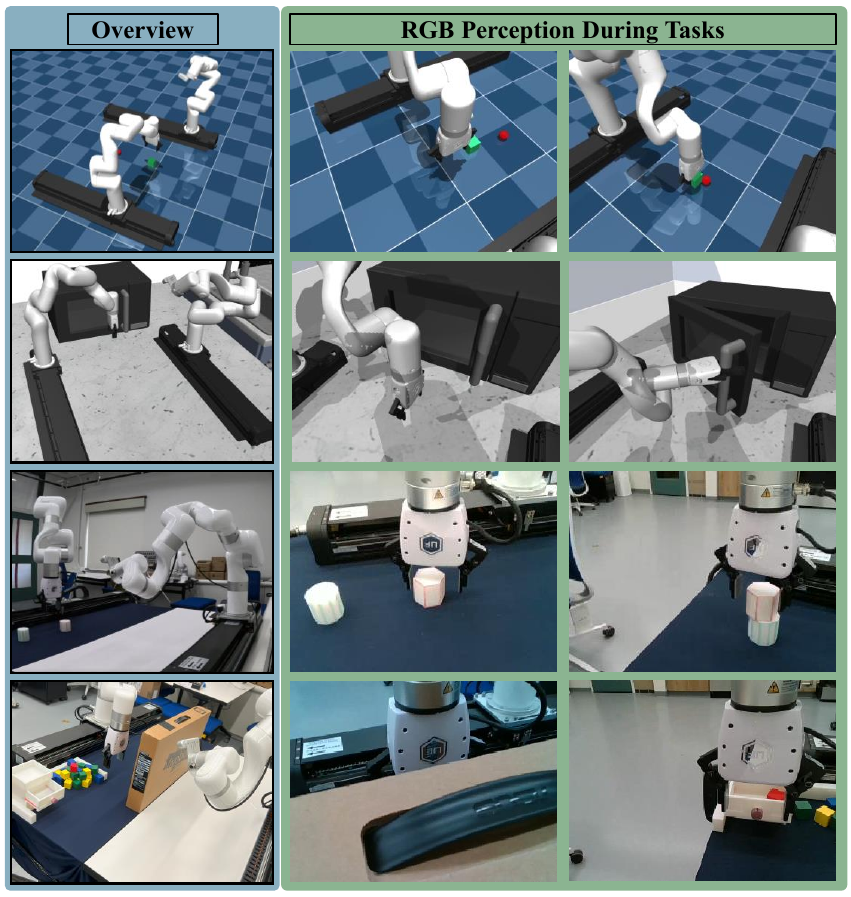}
\caption{Illustration of all simulated and real-world tasks, including an overview and the ego RGB perception of the system for each task. The first row depicts the simulated pick-and-place scenario. The second row shows the simulated microwave scenario. The third row presents the real-world block stacking scenario. The fourth row illustrates the real-world drawer interaction scenario.}
\label{fig:task_illustration}
\vspace{-0.6cm}
\end{figure}

\subsection{Implementation}
To demonstrate the effectiveness of our proposed method, we compared diffusion policies trained within different state-action spaces in various simulated and real-world tasks with demonstrations collected from teleoperation and the same hyperparameters as \cite{chi_diffusion_policy}. 
These baselines include the End-Effector Quaternion space ($\mathcal{E}_{\text{Q}}$), which aligns with the approach used in concurrent work on active perception and manipulation \cite{xiong2025vision}, allowing for a direct comparison.

We used PyTorch \cite{pytorch} as the machine learning framework for training the policy. All IK optimizations are performed using SLSQP \cite{kraft_slsqp} in NLopt \cite{NLopt}. Both training and inference were performed using an AMD PRO 5975WX CPU, dual 4090 GPUs, and 128GB RAM. The primary metrics throughout all evaluations were task success rate and the mean and standard deviation of task completion duration.

\begin{table}[htbp]
    \centering
    \renewcommand\arraystretch{1.2}
    \caption{Evaluation Results with Different State-Action Spaces \label{tab:eval_results}}
    \begin{tabular}{cl|rrr}
    \hline 
     \tabincell{c}{Task \\ (Training Epochs)} & Space &  \tabincell{c}{Mean \\ Duration} &  \tabincell{c}{Std \\ Duration} & \tabincell{c}{Success \\ Rate} \\
    \hline
    \multirow{6}*{\tabincell{c}{Simulated \\ Pick and Place \\ (200 Epochs)}} 
                       & $\mathcal{C}$ & N/A & N/A & 0.00 \\
                       & $\mathcal{E}_{\text{E}}$ & 340.52 stp & 93.69 stp &  0.84 \\
                       & $\mathcal{E}_{\text{Q}}$ & 410.76 stp & 96.06 stp & 0.46 \\
                       & $\mathcal{E}_{\text{AA}}$ & 480.18 stp & 52.47 stp & 0.13 \\
                       & $\mathcal{L}_{\text{E}}$ & \textbf{303.81 stp} &  69.38 stp &  \textbf{0.92} \\
                       & $\mathcal{L}_{\text{Q}}$ & 381.18 stp  & 95.24 stp & 0.56  \\
                       & $\mathcal{L}_{\text{AA}}$ & 498.55 stp & \textbf{13.47 stp} & 0.21 \\ 
    \hline
    \multirow{6}*{\tabincell{c}{Simulated \\ Microwave \\ (10 Epochs)}} 
                       & $\mathcal{C}$ & 446.98 stp & 93.83 stp & 0.30 \\
                       & $\mathcal{E}_{\text{E}}$ & 360.41 stp & 84.62 stp &  0.81 \\
                       & $\mathcal{E}_{\text{Q}}$ & 435.94 stp  & 100.77 stp & 0.34  \\
                       & $\mathcal{E}_{\text{AA}}$ & 468.66 stp & 108.99 stp & 0.09 \\ 
                       & $\mathcal{L}_{\text{E}}$ & \textbf{300.19 stp} &  \textbf{29.58 stp} &  \textbf{1.00} \\
                       & $\mathcal{L}_{\text{Q}}$ & 327.55 stp & 71.82 stp & 0.91 \\ 
                       & $\mathcal{L}_{\text{AA}}$ & 417.00 stp & 115.70 stp & 0.41 \\ 
    \hline
    \multirow{6}*{\tabincell{c}{Real-World \\ Block Stacking \\ (300 Epochs)}} 
                       & $\mathcal{C}$ & N/A & N/A & 0.00 \\
                       & $\mathcal{E}_{\text{E}}$ & 28.75 sec & 4.10 sec & 0.80 \\
                       & $\mathcal{E}_{\text{Q}}$ & 31.04 sec & 3.74 sec & 0.73 \\
                       & $\mathcal{E}_{\text{AA}}$ & 35.21 sec & 2.85 sec & 0.23 \\ 
                       & $\mathcal{L}_{\text{E}}$ & \textbf{25.65 sec} &  3.18 sec &  \textbf{1.00} \\
                       & $\mathcal{L}_{\text{Q}}$ & 29.12 sec & \textbf{2.78 sec} & 0.83 \\
                       & $\mathcal{L}_{\text{AA}}$ & 34.90 sec & 4.03 sec & 0.47 \\ 
    \hline
    \multirow{6}*{\tabincell{c}{Real-World \\ Drawer Interaction \\ (300 Epochs)}} 
                       & $\mathcal{C}$ & N/A & N/A & 0.00 \\
                       & $\mathcal{E}_{\text{E}}$ & 58.99 sec & 9.48 sec & 0.80 \\
                       & $\mathcal{E}_{\text{Q}}$ & 61.37 sec & 10.37 sec & 0.77 \\
                       & $\mathcal{E}_{\text{AA}}$ & 59.09 sec & 10.06 sec & 0.80 \\ 
                       & $\mathcal{L}_{\text{E}}$ & 56.86 sec & \textbf{6.63 sec} & 0.86 \\
                       & $\mathcal{L}_{\text{Q}}$ & \textbf{52.34 sec} & 8.97 sec & \textbf{0.87} \\
                       & $\mathcal{L}_{\text{AA}}$ & 56.80 sec & 7.58 sec & 0.83 \\ 
    \hline
    \label{eval_results}
    \end{tabular}
    \vspace{-0.6cm}
\end{table}


\subsection{Simulation Experiments}
\label{sec:simulation_experiments}
The simulation environments for dual-arm active perception and manipulation, developed using MuJoCo \cite{mujoco}, include a pick-and-place scenario and a microwave scenario. In the pick-and-place scenario, the agent must pick up a green cube and place it on a red goal position, with the cube randomly generated within a 15 cm radius around the goal. This scenario is shown in the first row of Fig. \ref{fig:task_illustration}. The microwave scenario, depicted in the second row of Fig. \ref{fig:task_illustration}, requires the agent to open the microwave door by at least 0.5 radians. Results for all simulation experiments were calculated from 100 inference rollouts. The pick-and-place policy was trained with 282 rollouts, and the microwave policy with 151 rollouts.

The simulation results in Tab. \ref{tab:eval_results} (row 1-2) show that our framework, which leverages the look-at IK, consistently and significantly outperforms the baselines. Diffusion policies trained in the look-at spaces achieve higher success rates—up to 4.5 times greater than their end-effector counterparts—and exhibit more stable performance with lower variance in task duration.  

This performance gain, however, cannot be attributed solely to dimensionality reduction. We observe two counter-intuitive results: first, there are instances where policies in lower-dimensional spaces underperform those in higher-dimensional ones---for example, the policy in $\mathcal{L}_{\text{AA}}$ (11-dim) underperforms the one in $\mathcal{E}_{\text{E}}$ (13-dim)---contradicting the common assumption that reducing dimensionality simplifies the learning problem and improves performance; second, representations using Euler angles consistently outperform theoretically superior quaternions and axis-angles, despite their known issues with singularities and discontinuities~\cite{spong_textbook, hashim_euler}.
These findings strongly suggest that a different, underlying factor is at play, motivating our frequency analysis in \S\ref{sec:frequency_analysis}.

\begin{figure}
\centering
\includegraphics[width=\columnwidth]{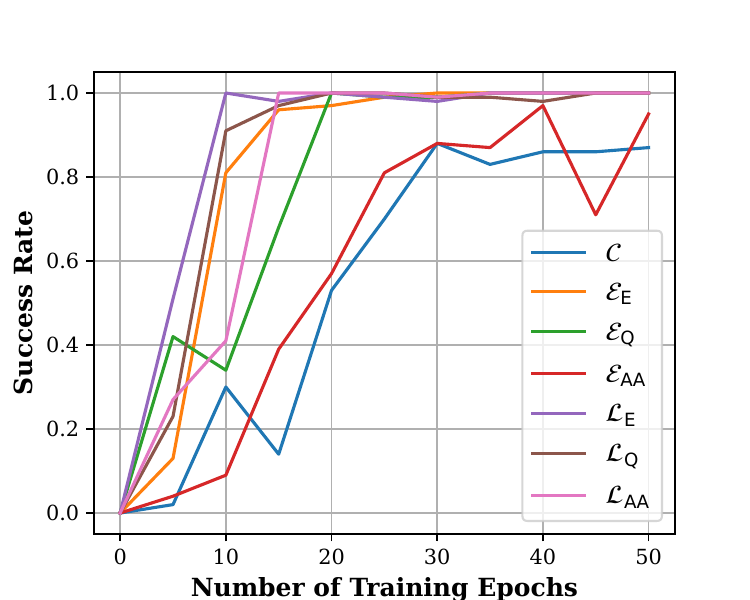}
\caption{The success rates of policies for the microwave task with different state-action spaces over 50 training epochs.}
\label{fig:microwave_training_epochs}
\vspace{-0.6cm}
\end{figure}

The microwave task was easier to learn than the pick-and-place task. For example, the policy trained with $\mathcal{L}_{\text{E}}$ reached a success rate of 1.00 after 10 training epochs, while the pick-and-place task reached 0.92 after 200 epochs despite having a larger demonstration dataset. Given the ease of the microwave task, we explored whether additional training epochs could help policies from C-Space and with end-effector IK achieve comparable performance. As shown in Fig. \ref{fig:microwave_training_epochs}, increasing training epochs led to success rates of baseline policies becoming more comparable, indicating that performance differences are likely due to better training efficiency of our proposed framework. This also indicates that diffusion policies can capture the data pattern easier with look-at IK compared to baselines.

\subsection{Real-World Experiments}
\label{sec:real_world_experiments}
We conducted real-world experiments in two scenarios: block stacking using 3D-printed objects from \cite{lee_block_stacking} and drawer interaction with a 3D-printed drawer from \cite{heo2023furniturebench}. The block stacking task involved non-cuboid objects, requiring precise gripper orientation for successful grasping, while the drawer interaction task required the agent to get rid of visual occlusion from a cardboard box, place a red target cube into the drawer, and close it. The block stacking policy was trained with 201 rollouts, and the drawer interaction policy with 67 rollouts. Each task was evaluated with 30 inference rollouts. As shown in Tab. \ref{tab:eval_results} (rows 3-4), policies learned with our look-at IK again yields substantial improvements in success rate and performance stability across both real-world tasks. The results also reinforce that dimensionality is not the primary performance driver, as policies trained in quaternion and axis-angle spaces—which have the same dimensionality—show markedly different success rates.

\subsection{Frequency Analysis}
\label{sec:frequency_analysis}

\begin{figure}[ht]
\centering
\includegraphics[width=\columnwidth]{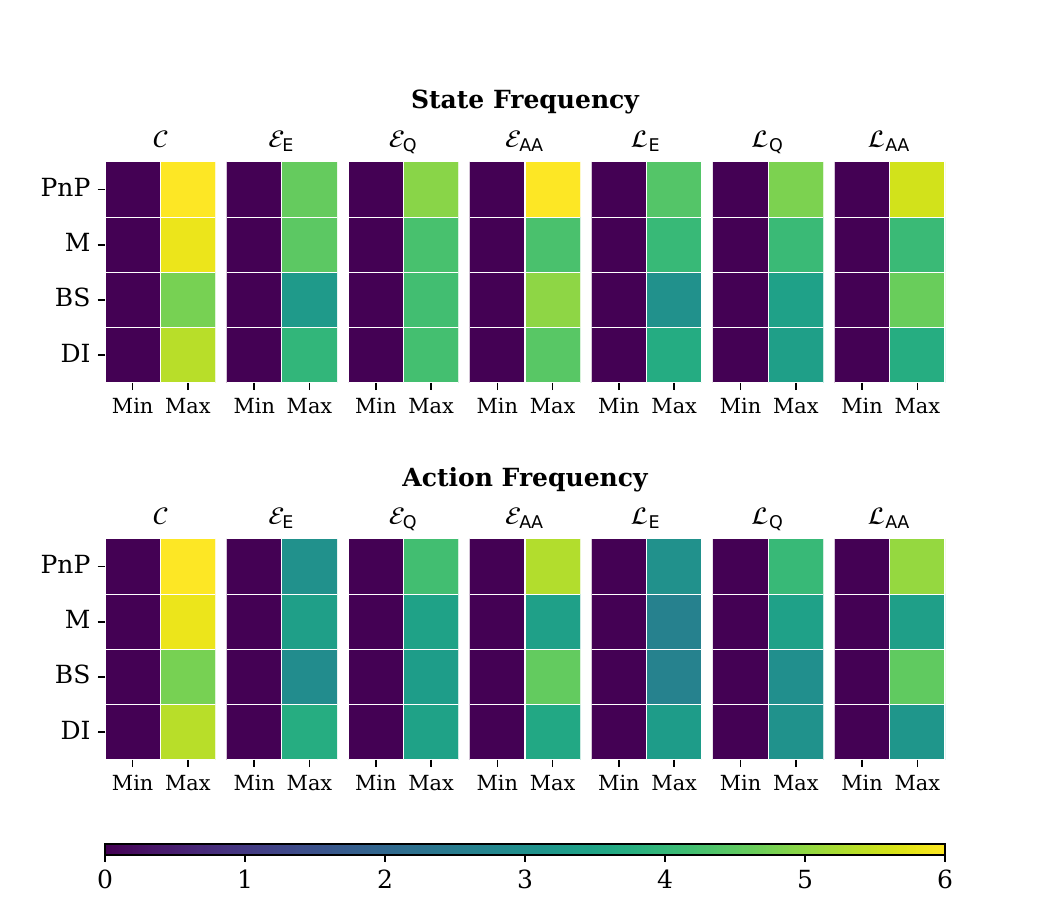}
\caption{Frequency heatmaps of concatenated state and action arrays from demonstration datasets across different evaluation scenarios (PnP for pick-and-place, M for microwave, BS for block stacking, DI for drawer interaction). Min and Max ticks indicate the lowest and highest frequency components. Color intensity denotes the logarithmically scaled magnitude of the frequency components.}
\label{fig:frequency_analysis}
\vspace{-0.35cm}
\end{figure}

To further investigate the performance improvement when training diffusion policies with our proposed look-at IK, we conducted a frequency analysis inspired by recent work on the interplay between frequency and CNNs \cite{wang_high_freq_cnn}. We performed a Fast Fourier Transform to convert the time-domain signals into the frequency domain and calculated the logarithm of the magnitudes of the Fourier coefficients:
$$
\mathbf{X}_k = \frac{1}{M} \sum_{m=1}^{M} \log\left(1 + \left| \sum_{n=0}^{N-1} \mathbf{x}_{m,n} \cdot e^{-i \frac{2\pi k n}{N}} \right|\right)
$$
where $\mathbf{X}_k$ is the $k_{th}$ component of the frequency-domain representation, $\mathbf{x}_{m,n}$ is the $n_{th}$ sample of the $m_{th}$ element in corresponding space with dimensionality of $M$ in a dataset with length of $N$. 

The results, shown in Fig. \ref{fig:frequency_analysis}, reveal that policy success rate is more strongly correlated with the magnitude of HFC than with space dimensionality. Notably, our proposed look-at spaces consistently exhibit lower HFC magnitudes across all tasks when compared to the baseline C-Space and end-effector spaces. 
This analysis also explains the surprising effectiveness of Euler angles, as they consistently produce lower HFC than other rotation representations, while axis-angle representations exhibit the highest.
This finding aligns with machine learning literature showing that common diffusion backbones, which rely on residual connections, are less effective at learning from HFC \cite{wang_high_freq_cnn, ronneberger_unet}. Therefore, by creating a state-action space with naturally lower HFC, our framework makes the learning problem more tractable for the diffusion policy, leading to the superior performance and efficiency observed in our experiments.

\section{Discussion}
\label{sec:discussion}

In this work, we introduced a framework that integrates a diffusion policy with a novel Look-At IK solver to address the challenges of manipulation with active perception. While concurrent work has approached this problem with standard end-effector IK \cite{xiong2025vision}, our method infuses a structural prior into the system. By automating viewpoint orientation, our framework creates a simplified space that is easier to learn for diffusion policies.

Beyond the immediate performance gains, our analysis provides a deeper insight by linking this success to the frequency characteristics of the space. We provide empirical evidence that the difficulty faced by standard approaches is strongly correlated with HFC, a factor not yet widely considered in the robotics community for this type of task. We hope this analysis encourages the community to consider robot learning problems through a new lens.

While our Look-At IK solver effectively mitigates the HFC issue for active perception, it is a task-specific solution. The fundamental challenge—that common diffusion backbones struggle to learn HFC—remains a general problem for many robotics tasks. Therefore, a critical direction for future work is to design new diffusion backbones that can capture HFC more effectively.





\bibliographystyle{plainnat}

\bibliography{refs}


\end{document}